\documentclass{article}

\usepackage[english]{babel}
\date{}

\usepackage[letterpaper,top=2cm,bottom=2cm,left=3cm,right=3cm,marginparwidth=1.75cm]{geometry}

\usepackage{amsmath}
\usepackage{graphicx}
\usepackage[table,xcdraw]{xcolor}
\usepackage[colorlinks=true, allcolors=blue]{hyperref}
\usepackage[most]{tcolorbox}
\usepackage{cite}
\usepackage[T1]{fontenc}
\usepackage{multirow}
\usepackage{authblk}

\title{ \textsc{Reddit-Impacts}: A Named Entity Recognition Dataset for Analyzing Clinical and Social Effects of Substance Use Derived from Social Media}

\author[1]{Yao Ge}
\author[1]{Sudeshna Das}
\author[2]{Karen O'Connor}
\author[3]{Mohammed Ali Al-Garadi}
\author[4]{Graciela Gonzalez-Hernandez}
\author[1,5,*]{Abeed Sarker}

\affil[1]{Department of Biomedical Informatics, School of Medicine, Emory University, Atlanta, GA}
\affil[2]{DBEI, The Perelman School of Medicine, University of Pennsylvania, Philadelphia, PA}
\affil[3]{Department of Biomedical Informatics, 
Vanderbilt University Medical Center, Nashville, TN}
\affil[4]{Department of Computational Biomedicine, Cedars-Sinai Medical Center, West Hollywood, CA}
\affil[5]{Department of Biomedical Engineering, Georgia Institute of Technology and Emory University, Atlanta, GA}

\begin{document}
\maketitle

\begin{abstract}
Substance use disorders (SUDs) are a growing concern globally, necessitating enhanced understanding of the problem and its trends through data-driven research. Social media are unique and important sources of information about SUDs, particularly since the data in such sources are often generated by people with lived experiences. In this paper, we introduce \textsc{\textsc{Reddit-Impacts}}, a challenging Named Entity Recognition (NER) dataset curated from subreddits dedicated to discussions on prescription and illicit opioids, as well as medications for opioid use disorder. The dataset specifically concentrates on the lesser-studied, yet critically important, aspects of substance use—its clinical and social impacts. We collected data from chosen subreddits using the publicly available Application Programming Interface for Reddit. We manually annotated text spans representing clinical and social impacts reported by people who also reported personal nonmedical use of substances including but not limited to opioids, stimulants and benzodiazepines. Our objective is to create a resource that can enable the development of systems that can automatically detect clinical and social impacts of substance use from text-based social media data. The successful development of such systems may enable us to better understand how nonmedical use of substances affects individual health and societal dynamics, aiding the development of effective public health strategies.  

In \textsc{\textsc{Reddit-Impacts}} dataset, we have a total of 1,380 posts, among which 23\% contain words or phrases annotated as clinical or social impacts. Specifically, 246 posts include entities annotated as having clinical impacts, and 72 posts are related to social impacts.

In addition to creating the annotated data set, we applied several machine learning models to establish baseline performances. Specifically, we experimented with transformer models like BERT, and RoBERTa, one few-shot learning model DANN by leveraging the full training dataset, and GPT-3.5 by using one-shot learning, for automatic NER of clinical and social impacts. The dataset has been made available through the 2024 SMM4H shared tasks.



\end{abstract}

\section{Introduction}

Substance use disorders represent a critical challenge in public health, with both clinical and social consequences impacting individuals and communities worldwide~\cite{lander2013impact,luciana2018adolescent}. The pervasive nature of substance use, encompassing both prescription and illicit drugs, necessitates a deeper understanding of its impacts to inform more effective interventions and preventative measures~\cite{clark2016impact,degenhardt2012extent}. This study introduces the \textsc{\textsc{Reddit-Impacts}} dataset, a unique corpus derived from Reddit, a platform known for its rich, anonymized discussions among diverse groups, including individuals who use drugs~\cite{guo2023generalizable, sarker2022social, rhidenour2022mediating}. The dataset includes posts from 14 opioid-related subreddits, capturing a broad spectrum of experiences and discussions related to substance use.

Our research specifically focuses on the clinical and social impacts of nonmedical substance use. These impacts are critical yet under-represented in the available data, making them an ideal focus for applying few-shot learning techniques to improve named entity recognition (NER) tasks. The clinical impacts encompass the direct effects on an individual’s health, while social impacts involve the broader consequences on relationships, communities, and societal structures. While there is an abundance of such information on Reddit, they are embedded in vast volumes of other unrelated information, making it extremely challenging to detect them automatically with high accuracy from naturally distributed data.

This paper details the creation of the \textsc{Reddit-Impacts} dataset, describes our annotation process, provides data statistics, and discusses the application of supervised learning algorithms aimed at enhancing the detection and classification of clinical and social impacts. We employed two models to benchmark the performance of our few-shot learning approach: one utilizing the full training dataset and another leveraging one-shot capabilities of large language models (LLMs) like Generative Pre-trained Transformer (GPT~\cite{radford2018improving, brown2020language}). These models provide baseline performance metrics that can be used for comparative evaluation of future systems customized for this complex NER task. In addition, this study explores innovative methodologies in applying few-shot learning to improve data annotation efficiency and model performance in detecting nuanced entity types like clinical and social impacts.



\section{Methods}

This study was considered to be exempt (category 4; publicly available data) by the Institutional Review Board of Emory University. The overall study can be divided into 4 steps: (1) data collection, (2) manual annotation, (3) creation of the \textsc{Reddit-Impacts} dataset and (4) NER.

\subsection{Data collection}
Reddit is popular in the broader community of people who use drugs as it offers anonymity, and Reddit has seen rapid growth in its user base over the last several years. Reddit communities have also been found to serve as a means of social support for people who use drugs. We chose Reddit over other social networks or web-based forums such as Twitter, Bluelight, and Discord for several reasons. While all these sources contain information about substance use, the substance use community of Reddit is much larger and has been extensively used in peer-reviewed research related to substance use and emerging substance use trends. Additionally, Reddit threads are also heavily moderated, and posts must follow community-specific rules. Consequently, while these rules restrict some types of information from being posted, they also ensure that the data are reflective of the topical areas and the volume of spam, posts from bots, or irrelevant content is thereby lower. The existence of standard application programming interfaces (APIs) also makes data collection from Reddit relatively straightforward.

To identify potential Redditors (Reddit subscribers) who self-report opioid usage on Reddit, we identified 14 opioid-related subreddits spanning discussions on prescription and illicit opioids, and collected all retrievable posts using the Python-Reddit API Wrapper for Reddit (PRAW).\footnote{\url{https://praw.readthedocs.io/en/latest/}}

The choice of these subreddits was based on their topical relevance and high levels of community discussion and engagement. Collection of data from these subreddits was not keyword-based. Instead, the API allowed the retrieval of all publicly posted threads and the associated comments. After retrieving all available posts of the 47,327 Redditors who had posted on the selected sub-reddits, we selected a random sample of these Redditors (N=13,812) and collected each of their past public posts across all subreddits (i.e., their longitudinal timelines), between November 2006 (corresponding to the earliest post available) and March 2019 (corresponding to the last date of data collection). 

\subsection{Annotation}

From the 13,812 public timelines we collected, we randomly selected 40 Redditors' timelines (i.e., all their posts in different subreddits) for manual review and annotation. This process finally yielded 26,126 posts for annotation. The annotation process was iterative and involved several steps. The posts were manually analysed to develop the annotation guidelines, and then preliminary rounds of annotation were performed. We then discussed the disagreements, and updated the annotation guidelines for further clarity, and the final annotation was performed on a total of 91,601 sentences (2,500,489 tokens). 

Due to the complexity of the annotation task, involving many entity types, and large numbers of posts that contained no entities at all, rather than annotating separately and then computing inter-annotator agreement, the data was first annotated by the lead annotator (KO) based on annotation guidelines and reviewed by two members of the study team. Following the annotation of all posts by two subscribers, the annotations were reviewed by the full team, disagreements were resolved via discussion and the annotation guideline was updated. Subsequent annotations were carried out in the same manner, adhering to the annotation guideline.  All disagreements were resolved via discussion. 

\begin{figure}[h!]
\centering
\includegraphics[width=0.6
\columnwidth]{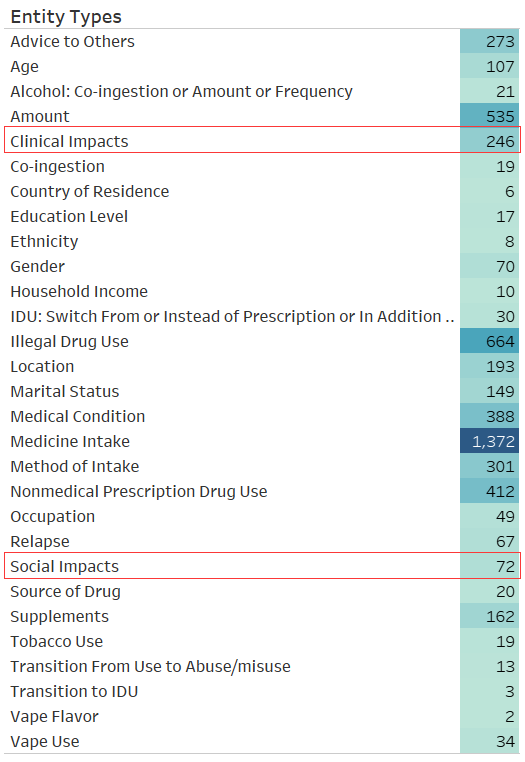}   
\caption{Entity types and the number of posts in each entity type.}  
\label{Entity_Types}
\end{figure}

Based on the annotation guidelines we annotated lexical expressions in posts into 30 entity types that are independent of each other. Among them, 10 entity types belong to the basic personal information category, such as Age, Gender, Marital status, Location, Income, etc. 20 entity types related to medication information, such as Medicine intake, Illegal drug use, NMPDU, Method of intake, etc. Figure \ref{Entity_Types} shows all 30 entity types and their statistics in the annotated dataset. 

The annotation process of our extensive dataset highlighted the prevalence of readily identifiable concepts such as medicine intake and illegal drug use. However, it also revealed that instances of clinical and social impacts---central to our study---are notably scarce. This scarcity poses significant challenges for research, as these impacts are crucial for understanding the broader consequences of nonmedical substance use on individual health and societal dynamics. To address these challenges and align with our objective of developing more effective public health strategies, we have concentrated our efforts on these two underrepresented entity types, thereby creating the specialized \textsc{Reddit-Impacts} dataset. This focused approach aims to enhance our ability to detect and study these rare but critical impacts in the discourse surrounding substance use.

\subsection{\textsc{Reddit-Impacts} Dataset}

\begin{table}[]
\centering
\resizebox{\textwidth}{!}{%
\begin{tabular}{lllll}
\hline
\textbf{Datasets} &
  \textbf{Entity Types} &
  \textbf{Training Size} &
  \textbf{Test Size} &
  \textbf{Entities} \\ \hline
\multirow{2}{*}{\textbf{\textsc{Reddit-Impacts}}} &
  \multirow{2}{*}{\begin{tabular}[c]{@{}l@{}}Clinical Impacts, \\ Social Impacts\end{tabular}} &
  30k tokens &
  6k tokens &
  0.2k tokens \\ \cline{3-5} 
 &
   &
  1,102 posts &
  278 posts &
  318 posts \\ \hline
\end{tabular}%
}
\caption{Statistics of \textsc{Reddit-Impacts} dataset, including training and test sizes, the number of entity types and the number of entities in the dataset.}
\label{tab:statistics}
\end{table}
\begin{table}[h!]
\centering
\begin{tabular}{lllll}
\hline
\textbf{Index} & \textbf{Span} & \textbf{Token} & \textbf{Entity or not} & \textbf{Label}   \\ \hline
85-1           & 13055-13057   & In             & \_                     & \_               \\ \hline
85-2           & 13058-13060   & PA             & \_                     & \_               \\ \hline
85-3           & 13061-13063   & at             & *                      & Clinical Impacts \\ \hline
85-4           & 13064-13065   & a              & *                      & Clinical Impacts \\ \hline
85-5           & 13066-13068   & 28             & *                      & Clinical Impacts \\ \hline
85-6           & 13069-13072   & day            & *                      & Clinical Impacts \\ \hline
85-7           & 13073-13078   & detox          & *                      & Clinical Impacts \\ \hline
85-8           & 13078-13079   & /              & *                      & Clinical Impacts \\ \hline
85-9           & 13079-13084   & rehab          & *                      & Clinical Impacts \\ \hline
85-10          & 13085-13089   & they           & \_                     & \_               \\ \hline
85-11          & 13090-13094   & used           & \_                     & \_               \\ \hline
85-12          & 13095-13104   & methadone      & \_                     & \_               \\ \hline
85-13          & 13105-13107   & to             & \_                     & \_               \\ \hline
85-14          & 13108-13111   & get            & \_                     & \_               \\ \hline
85-15          & 13112-13114   & me             & \_                     & \_               \\ \hline
85-16          & 13115-13118   & off            & \_                     & \_               \\ \hline
85-17          & 13119-13121   & of             & \_                     & \_               \\ \hline
85-18          & 13122-13126   & bupe           & \_                     & \_               \\ \hline
85-19          & 13126-13127   & .              & \_                     & \_               \\ \hline
\end{tabular}
\caption{A sample post "In PA at a 28 day detox / rehab they used methadone to get me off of bupe." with index, spans, tokens and corresponding labels.}
\label{tab:data_example}
\end{table}
From the total of 26,126 posts, only 318 posts (approximately 1.22\%) were annotated as having clinical or social impacts. This extremely low occurrence rate underscores the sparsity of relevant data within the larger dataset. Due to the vast size and sparse nature of the original dataset, we opted to randomly select a subset of 1,380 posts for our experiments. We divided the annotated data into 3 sets: 60\% for training, 20\% for validation, and 20\% for testing/evaluation. In summary, \textsc{Reddit-Impacts} comprises 843 posts for training, 259 for validation, and 278 for testing. 

This approach not only made the data more manageable but also ensured a focused analysis on the most relevant instances. By narrowing our dataset, we could intensify our efforts on enhancing the detection and classification of these rare but significant entities. This refined dataset formation was pivotal for our experiments and subsequent release of the \textsc{Reddit-Impacts} dataset for the SMM4H 2024 shared task, aiming to provide a resource that is both concentrated and rich in the entities of interest---clinical impacts and social impacts. The number of instances of our \textsc{Reddit-Impacts} dataset are also shown in Table \ref{tab:statistics}. In addition, table \ref{tab:data_example} presents an example of posts and their labels.

\subsection{Named Entity Recognition}
\subsubsection{Models}
Transformer-based approaches that use large pre-trained language models achieve state-of-the-art F$_1$-scores for NER tasks when large annotated data available. Due to the sparsity of annotated samples in the \textsc{Reddit-Impacts} dataset, we choose to fine-tune and evaluate two popular transformer-based models as the baseline: BERT~\cite{devlin:2019} and RoBERTa~\cite{liu:2019}. Building on prior research in few-shot learning, we also report performances for DANN (Data Augmentation with Nearest Neighbor classifier)~\cite{ge2024data}, which demonstrated promising performance in few-shot scenarios. 

Given the remarkable success of LLMs in few-shot learning scenarios, we also 
explore the viability of employing GPT-3.5 for the extraction of named entities in a one-shot setting (by providing one example of input data in the prompt). This evaluation aims to provide insights into the performance of GPT-3.5, further enriching our benchmarking of this dataset. 

The following is an outline of the models we used:

\textbf{1. BERT~\cite{devlin:2019}:} A foundational Transformer-based model, widely recognized for its pre-training on a large corpus of text from books and Wikipedia.

\textbf{2. RoBERTa~\cite{liu:2019}:} Transformer-based model popular for its training on big batches and long sequences.

\textbf{3. DANN~\cite{ge2024data}:} A few-shot learning method for NER that uses a data augmentation module combined with a nearest neighbor classifier to solve data sparsity problems. 

\textbf{4. GPT-3.5:} An advanced iteration of the Generative Pre-trained Transformer series, known for its enhanced language understanding and generation capabilities, trained on a diverse range of internet text.

\subsubsection{Evaluation Metrics}

We compared the performances of the models based on the micro-averaged F$_1$-score for clinical impacts and social impacts. We focused our evaluation on these two entity types since that is our class of interest. We report overall entity-level relaxed F$_1$-score, entity-level strict F$_1$-score, and token-level F$_1$-score on these two entity types. For entity-level relaxed F$_1$-scores, we use SemEval guidelines~\footnote{\url{https://www.davidsbatista.net/blog/2018/05/09/Named_Entity_Evaluation/}} to calculate 
partial matches between the predictions and gold-standard annotations.

\section{Results and Discussions}
\subsection{Data and annotation}

23\% of the posts in the dataset contain words or phrases marked as clinical impact or social impact, with 184 entities annotated as clinical impacts and 67 entities as social impacts. 

\subsection{Performance on NER Task}
\begin{table}[h!]
\begin{tabular}{lllll}
\hline
\textbf{Model} &
  {\color[HTML]{212121} \textbf{Training Size}} &
  {\color[HTML]{212121} \textbf{\begin{tabular}[c]{@{}l@{}}Entity-level\\ Relaxed F1-Score\end{tabular}}} &
  {\color[HTML]{212121} \textbf{\begin{tabular}[c]{@{}l@{}}Entity-level\\ Strict F1-Score\end{tabular}}} &
  {\color[HTML]{212121} \textbf{\begin{tabular}[c]{@{}l@{}}Token-level \\ F1-Score\end{tabular}}} \\ \hline
BERT    & Full training data & 0.0   & 0.0   & 0.0   \\ \hline
RoBERTa & Full training data & 0.0   & 0.0   & 0.0   \\ \hline
DANN    & Full training data & \textbf{54.36}  & \textbf{32.62} & \textbf{50.79}  \\ \hline
GPT-3.5 & One-shot           & \textbf{16.73} & \textbf{10.98} & \textbf{26.10} \\ \hline
\end{tabular}
\caption{Performance on baseline models, including training size we used, entity-level relaxed F$_1$-score, entity-level strict F$_1$-score, and token-level F$_1$-socre.}
\label{tab:results}
\end{table}
Table \ref{tab:results} presents the results of our automatic NER experiments. The table shows the overall relaxed F$_1$-score, strict F$_1$-score, and token-level F$_1$-score on two entity types: clinical impacts and social impacts. The DANN model achieved the highest F$_1$-score among all the models when the entire training data was used for training. 
We found that BERT and RoBERTa are unable to identify clinical and social impacts in the dataset despite using the full training data for fine-tuning.

In the few-shot settings, GPT-3.5 tends to perform well on the dataset, even though the evaluation was carried out in a one-shot setting. This demonstrated the significantly higher accuracy of LLMs such as GPT-3.5 in few-shot settings over fine-tuned pre-trained language models (PLMs) for entity extraction in clinical text. 

\section{Conclusions}
Our annotation effort highlighted that information about the clinical and social impacts of substance use are available on Reddit, albeit being sparse. Our experiments demonstrated the difficulty of automatically detecting the sparse clinical and social impact concepts via supervised machine learning, although the DANN model showed promising performance. There is room for improvement in this field, our future efforts will focus on leveraging advancements in large language models like Meta Llama 3~\footnote{\url{https://github.com/meta-llama/llama3}} to improve automatic NER performance and evaluate the applicability of our research in real-world settings.

\section*{Supplementary}

Listing 1: Example prompt we used for GPT-3.5

\begin{tcolorbox}[colframe=black]
You are a medical AI trained to identify and classify tokens into three categories: \textcolor{blue}{Clinical Impacts, Social Impacts, and Outside ('O')}. 'Clinical Impacts' refer to tokens describing the effects, consequences, or impacts of substance use on individual health or well-being, as defined in UMLS. 'Social Impacts' describe the societal, interpersonal, or community-level effects, also based on UMLS definitions. Any token not falling into these categories should be labeled as 'O'. 

For example, the sentence \textcolor{blue}{'I was a codeine addict.'} is tokenized and labeled as follows: 

\textcolor{blue}{['I', 'was', 'a', 'codeine', 'addict', '.']} with labels \textcolor{blue}{['O', 'O', 'O', 'Clinical Impacts', 'Clinical Impacts', 'O']}. 

Your task is to predict and return the label for each provided token, ensuring the number of output labels matches the number of input tokens exactly. 
The output format should be tokens with their labels: 
\textcolor{blue}{['I-O', 'was-O', 'a-O', 'codeine-Clinical Impacts', 'addict-Clinical Impacts', '.-O']}. 
\end{tcolorbox}

\bibliographystyle{unsrt}
\bibliography{main}

\end{document}